\pdfoutput=1
\documentclass[11pt]{article}
\usepackage[final]{ACL2023}
\usepackage{times}
\usepackage{latexsym}
\usepackage[T1]{fontenc}
\usepackage[utf8]{inputenc}
\usepackage{microtype}
\usepackage{inconsolata}
\usepackage{hyperref}
\usepackage{graphicx}
\title{Cross-lingual Emotion Detection through Large Language Models}
\author{Ram Mohan Rao Kadiyala \\
  University of Maryland, College Park \\
  \texttt{rkadiyal@terpmail.umd.edu}}

\begin{document}
\maketitle
\begin{abstract}
This paper presents a detailed system description of our entry which finished 1st with a large lead at WASSA 2024 Task 2, focused on cross-lingual emotion detection. We utilized a combination of large language models (LLMs) and their ensembles to effectively understand and categorize emotions across different languages. Our approach not only outperformed other submissions with a large margin, but also demonstrated the strength of integrating multiple models to enhance performance. Additionally, We conducted a thorough comparison of the benefits and limitations of each model used. An error analysis is included along with suggested areas for future improvement. This paper aims to offer a clear and comprehensive understanding of advanced techniques in emotion detection, making it accessible even to those new to the field.
\end{abstract}


\section{Introduction}
Emotion detection in texts across different languages is a challenging yet crucial task, especially in the context of global digital communication. The ability to accurately identify emotions in text, regardless of the language, can significantly enhance interactions in various domains such as customer service, social media monitoring, and mental health assessments. This paper introduces our approach to cross-lingual emotion detection, which was recently recognized as the top submission in WASSA 2024 Task 2 \citep{Maladry2024}. Our system leveraged the capabilities of several open source and proprietary Large Language Models (LLMs), including GPT-4 \citep{openai2024gpt4} and Claude-Opus \citep{opus2024} in a zero-shot configuration, as well as LLAMA-3-8B \citep{touvron2023llama}, Gemma-7B \citep{gemmateam2024gemma}, and Mistral-v2-7B \citep{jiang2023mistral}, which were fine-tuned. To assess the robustness and efficiency of these models, we conducted tests in both 4-bit and 16-bit precision. This varied precision testing helps in understanding the trade-offs between computational efficiency and model performance. Additionally, we compared the performance of our models against the top submission's \citep{patkar-etal-2023-adityapatkar} approach on a similar monolingual task from the previous years' shared task. Furthermore, we experimented with enhancing model performance by incorporating additional training data from previous editions of the shared task, specifically WASSA 2023 \citep{barriere-etal-2023-findings} and WASSA 2022 \citep{barriere-etal-2022-wassa} emotion classification task datasets.


\section{Dataset}
The dataset consisted of texts belonging to one the 5 languages - Dutch, English, French, Russian and Spanish annotated as one of the 6 classes - Anger, Fear, Love, Joy, Neutral and Sadness. The distribution of languages and each class in each of the datasets can be seen in \autoref{table:1} and \autoref{table:2}.

\begin{table}[!ht]
    \centering
    \begin{tabular}{|c|c|c|c|}
    \hline
    \textbf{Class $\downarrow$}  & \textbf{Train}       & \textbf{Dev}         & \textbf{Test}    \\
    \hline     
    \textbf{Anger}               & 1028                 & 129                  & 614              \\
    \textbf{Fear}                & 143                  & 14                   & 77               \\
    \textbf{Joy}                 & 1293                 & 102                  & 433              \\
    \textbf{Love}                & 579                  & 40                   & 190              \\
    \textbf{Neutral}             & 1397                 & 157                  & 916              \\
    \textbf{Sadness}             & 560                  & 58                   & 270              \\
    \hline
    \textbf{Total}               & 5000                 & 500                  & 2500             \\
    \hline
    \end{tabular}
    \caption{Class distribution in each dataset split}
    \label{table:1}
\end{table}

\begin{table}[!ht]
    \centering
    \begin{tabular}{|c|c|c|c|}
    \hline
    \textbf{Class $\downarrow$}  & \textbf{Train}       & \textbf{Dev}         & \textbf{Test}    \\
    \hline
    \textbf{English}             & 5000                 & 100                  & 500              \\
    \textbf{French}              & -                    & 100                  & 500              \\
    \textbf{Dutch}               & -                    & 100                  & 500              \\
    \textbf{Russian}             & -                    & 100                  & 500              \\
    \textbf{Spanish}             & -                    & 100                  & 500              \\
    \hline
    \textbf{Total}               & 5000                 & 500                  & 2500             \\
    \hline
    \end{tabular}
    \caption{Language distribution in each dataset split}
    \label{table:2}
\end{table}


\begin{figure*}[!ht]
    \centering
    \includegraphics[width=1\linewidth]{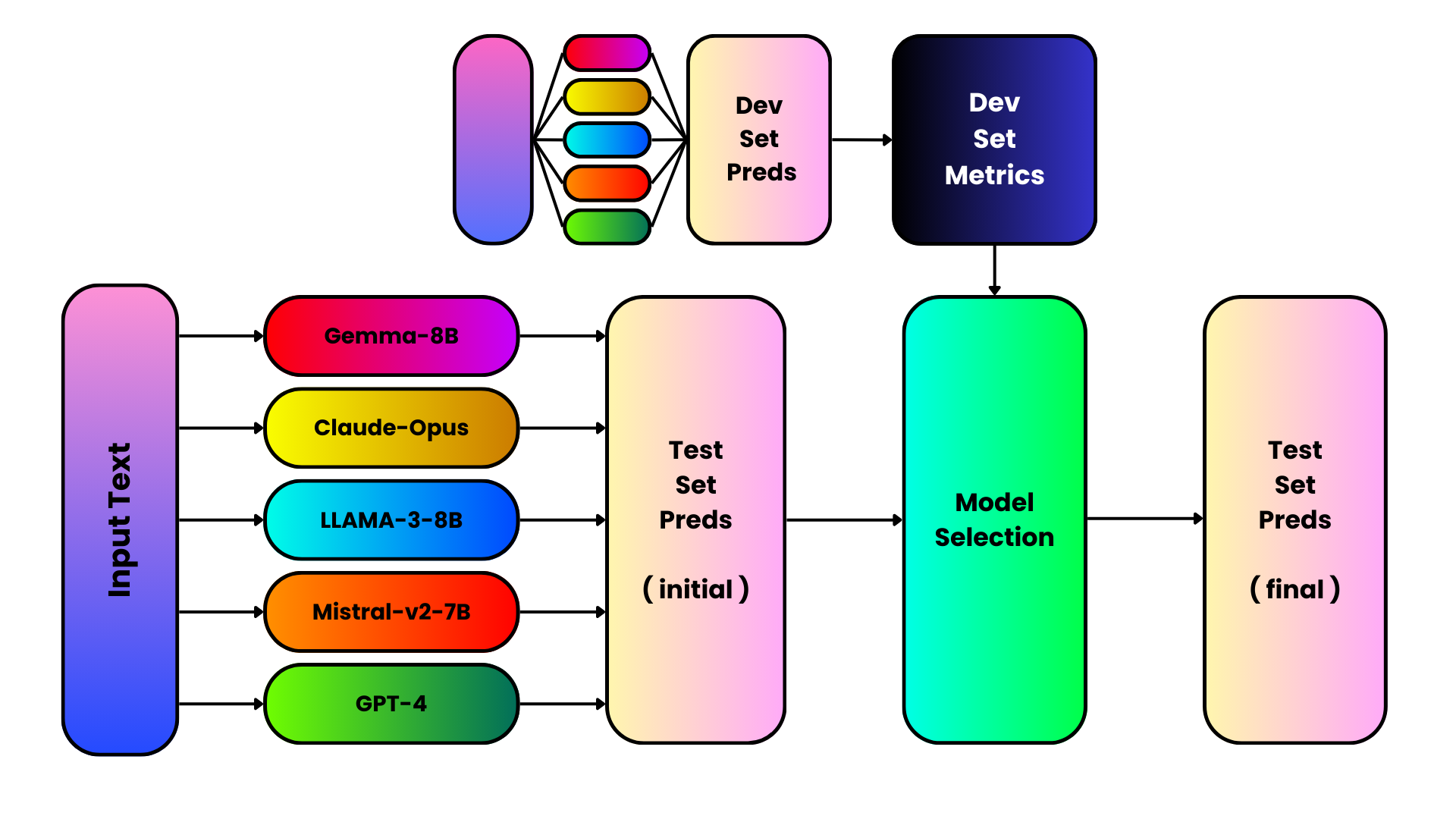}
    \caption{System Overview : Ensembles of LLMs}
    \label{figure:1}
\end{figure*}

\section{System Description}

The non-proprietary LLMs were fine-tuned over just the training dataset over 5 epochs with a learning rate of 0.0002 and weight decay of 0.01. The proprietary systems were tested with various prompt over the development set and the best performing prompt was used to make predictions over the test set. Additionally the previous year's benchmark was also tested alongside by replacing RoBERTa \citep{liu2019roberta} with XLM-RoBERTa \citep{conneau2020unsupervised}. Additionally other ensembles like majority vote, model selection based on features were also tested. The Code and Models are available over the GitHub repository\footnote{Code Used : \url{https://github.com/1024-m/ACL-2024-WASSA-EXALT}} and Huggingface\footnote{The finetuned LLAMA Model : \url{https://huggingface.co/1024m/EXALT-1A-LLAMA3-5A-16bit}} \footnote{The fine-tuned Mistral Model : \url{https://huggingface.co/1024m/EXALT-1A-MISTRAL-5A-16bit}} \footnote{The finetuned GEMMA Model : \url{https://huggingface.co/1024m/EXALT-1A-GEMMA-5A-16bit}}.  The primary metric was weighted F1 score, additionally Precision and Recall have also been observed.


\subsection{Results Comparison}
The results using each of the models on the development set by fine-tuning over 3 epochs on the training set can be seen in \autoref{table:3}. Other approaches like data augmentation using previous years' emotion detection datasets, translating dev and test sets to English before making predictions did not improve the metrics. No pre-processing steps have been used. The metrics on the Test set can be seen in \autoref{table:4}.

\begin{table}[!ht]
    \centering
    \begin{tabular}{|c|c|c|}
    \hline
    \textbf{Model $\downarrow$} & \textbf{Description} & \textbf{F1} \\
    \hline
    GPT-4            & Zero-shot   & \textbf{0.5616}  \\
    Claude-Opus      & Zero-shot   & 0.5581  \\
    LLaMa-3-8B       & Fine-tuned 3 epochs & 0.5474  \\
    Mistral-v2-7B    & Fine-tuned 3 epochs & 0.5466 \\
    Gemma-8B         & Fine-tuned 3 epochs & 0.5300  \\
    \hline     
    Xlm-R            & 10 epochs + SWA & 0.5392  \\
    \hline     
    \end{tabular}
    \caption{Performance of each model on Dev set}
    \label{table:3}
\end{table}

\begin{table*}[ht]
    \centering
    \begin{tabular}{|c|c|c|c|c|}
    \hline
    \textbf{Model $\downarrow$} & \textbf{Description} & \textbf{F1 score} \\
    \hline
    llama-3-8b       & fine-tuned , 5 epochs                               & 0.5931  \\
    llama-3-8b       & fine-tuned , test data translated , 5 epochs        & 0.5701  \\
    gemma-8b         & fine-tuned , 5 epochs                               & 0.5450  \\
    mistral-v2-7b    & fine-tuned , 5 epochs                               & 0.5915  \\
    gpt-4            & few-shot : one sample of each class                 & 0.5918  \\
    claude-opus      & zero-shot                                           & 0.5257  \\
    ensemble         & model selection based on weighted-f1 scores , 5 epochs each        & 0.5810  \\
    ensemble         & model selection based on macro-f1 scores , 5 epochs each           & 0.5977  \\
    ensemble         & model selection based on micro-f1 scores , 5 epochs each           & 0.5725  \\
    ensemble         & majority vote or model selection based on macro-f1 , 5 epochs each & \textbf{0.6295}  \\
    \hline     
    \end{tabular}
    \caption{Performance of each models / approaches on Test set}
    \label{table:4}
\end{table*}


\subsection{Error Analysis}
Each of the models had its own advantages and drawbacks likely due to the differences in the pre-training data used by each of the models. The performance of each of the models was observed separately on each of the languages over the development set, this can be seen in \autoref{table:5}. It can be seen that certain models performed better on some of the languages. This led to the conclusion that selecting an appropriate model based on language of the text to be classified might yield better results.

\begin{table*}[!ht]
    \centering
    \begin{tabular}{|c|c|c|c|c|c|c|}
    \hline
    \textbf{Language} & \textbf{Metric} & \hspace{0.4em}\textbf{GPT-4}\hspace{0.4em} & \textbf{GEMMA} & \textbf{Claude-Opus} & \textbf{Mistral-v2} & \textbf{LLAMA-3}\\
    \hline         
    English & Micro F1    & 0.610 & 0.650          & 0.580 & \textbf{0.680} & 0.610 \\
    English & Macro F1    & 0.443 & \textbf{0.594} & 0.470 & 0.590          & 0.481 \\
    English & Weighted F1 & 0.582 & 0.655          & 0.563 & \textbf{0.671} & 0.587 \\
    \hline
    Russian & Micro F1    & \textbf{0.620} & 0.550 & 0.570 & 0.590 & 0.610 \\
    Russian & Macro F1    & \textbf{0.506} & 0.425 & 0.454 & 0.434 & 0.457 \\
    Russian & Weighted F1 & \textbf{0.633} & 0.574 & 0.584 & 0.603 & 0.627 \\
    \hline     
    Spanish & Micro F1    & 0.670 & 0.700 & 0.630 & 0.740 & \textbf{0.770} \\
    Spanish & Macro F1    & 0.521 & 0.552 & 0.597 & 0.659 & \textbf{0.687} \\
    Spanish & Weighted F1 & 0.676 & 0.725 & 0.666 & 0.751 & \textbf{0.779} \\
    \hline     
    French  & Micro F1    & 0.590 & 0.610          & 0.610 & \textbf{0.630} & \textbf{0.630} \\
    French  & Macro F1    & 0.509 & 0.533          & 0.499 & \textbf{0.549} & 0.522          \\
    French  & Weighted F1 & 0.579 & \textbf{0.607} & 0.596 & 0.596          & 0.589          \\
    \hline     
    Dutch   & Micro F1    & \textbf{0.670} & 0.550 & 0.650          & 0.620 & 0.660 \\
    Dutch   & Macro F1    & \textbf{0.540} & 0.394 & \textbf{0.540} & 0.413 & 0.533 \\
    Dutch   & Weighted F1 & \textbf{0.657} & 0.576 & 0.636          & 0.610 & 0.642 \\
    \hline
    \end{tabular}
    \caption{Performance of each model on Dev set : by each Language and Metric}
    \label{table:5}
\end{table*}


\subsection{Our System}
Several approaches of using ensembles based on majority voting, model selection based on macro F1, micro F1 and the weighted F1 scores were tested. The best performing system uses a majority voting criteria from the 5 models used. In cases where consensus is not achieved i.e no clear majority, the output of the model with highest weighted F1 score was chosen as the final label.  


\subsection{Possible Extensions}
As seen in \autoref{table:5}, each of the models had their own advantages and disadvantages with varying performances on each language. It is likely that adding more models into the system and features like text length or utilizing different models for binary classification of whether the given text belongs to a class. This can be seen in \autoref{table:6} displaying varying effectiveness of each model in predicting each emotion. A viable approach would be predicting each emotion as a binary task and then using other methods in cases where none or more than one class ends up as true. The fine-tuned LLMs were loaded in 4bit precision and later fine tuned using LoRA \citep{hu2021lowrank} and tested in both 4bit precision and 16bit precision versions. The drop in performance in 4bit overall was minimal, however in many cases the predictions in 4bit ended up as correct while 16bit were incorrect. Another viable approach is to pick the top 2 likely class labels for each of the texts' predictions and using other methods to classify more effectively.

\begin{table*}[!ht]
    \centering
    \begin{tabular}{|c|c|c|c|c|c|}
    \hline
    \textbf{Class $\downarrow$}  & \hspace{1em}\textbf{GPT-4}\hspace{1em} & \textbf{GEMMA-8B} & \textbf{Claude-Opus} & \textbf{Mistral-v2-7B} & \textbf{LLAMA-3-8B} \\
    \hline
    Anger       & \textbf{0.75} & 0.69          & 0.71          & 0.72          & 0.74          \\
    Fear        & 0.26          & 0.27          & \textbf{0.42} & 0.30          & 0.40          \\
    Joy         & 0.62          & 0.59          & 0.61          & \textbf{0.67} & 0.65          \\
    Love        & 0.40          & 0.44          & 0.33          & \textbf{0.46} & 0.34          \\
    Neutral     & 0.69          & 0.73          & 0.67          & 0.74          & \textbf{0.75} \\
    Sadness     & 0.42          & \textbf{0.47} & 0.45          & 0.39          & 0.40          \\
    \hline
    \end{tabular}
    \caption{Performance of each model class wise : class F1 scores on Dev set}
    \label{table:6}
\end{table*}


\section{Conclusion}
It can be seen from \autoref{table:4} that ensemble models have achieved a significantly better result over direct approaches. However not all approaches have been tested due to limit on number of submissions. As seen in \autoref{table:5}, It can also be observed that from \autoref{table:6} that a similar trend was observed in using different models for each emotion detection too might aid in improving the performance further. As seen in \autoref{table:4} and \autoref{table:3}, further training is likely to improve the results as the dev set results of fine-tuned models were lower that the proprietary models when trained on 3 epochs, but when the same models were further tuned over 2 more epochs, they performed better than proprietary models. Most of the errors when using proprietary models were with the neutral class texts being classified incorrectly or other classes being classified as being neutral. While the fine-tuned models were able to learn to be able to distinguish texts as neutral or some other class in a better way as seen in \autoref{table:6}. The classes with lesser data samples as shown in \autoref{table:1} had significantly worse compared to other classes as seen in \autoref{table:6}. Techniques like Stochastic weight averaging (SWA) \citep{izmailov2019averaging} in this case only led to a minor improvement and techniques like augmentation using other datasets did not improve performance. It is likely that adding sufficient data for all classes can make the current proposed system better as enough correlation can be seen in training data amount from \autoref{table:1} and average performance of the discussed models on each of the classes from \autoref{table:6}.  The current proposed approach can be extended to other languages by testing performance on a small sample of that language to decide the extent of reliability of each model in making predictions over texts of that language. In case of using proprietary models the same prompt used for all texts, it is worth testing different prompts for texts of each language due to varying features of each language where one class might to higher number of false positives than other in a different language. Approaches like removal of stop words did not improve the performance. While using ensembles, texts completely in one language performed better that the texts where a portion of the text is in English and rest in a different language. These texts led to higher frequency of errors. The performance of proprietary models was a bit better on these kind of texts compared to the rest of the models tested probably due to larger model size and more code-mixed data in training. Other information like the specific prompts used on each of the LLMs, Prompt format for the fine-tuned LLMs used and other relevant plots are available in \autoref{Appendix}. 

\section*{Limitations}
Due to computational resource limitations, the models used (non-proprietary) were loaded in 4bit precision before being fine-tuned. It is likely that with higher precision usage of the models can yield better results. The models used (non-proprietary) were of the 7B or 8B variants. It is likely that larger variants may yield better results. The approaches might not be extendable to all languages as not all languages' data were covered in the pre-training data of the LLMs used in the current proposed system. Due to time limitations, not all LLMs were tested, especially some of the other proprietary LLMs which might perform better in one of languages in consideration.

\nocite{mohammad-etal-2018-semeval, chatterjee-etal-2019-semeval}
\bibliography{anthology,custom}

\begin{thebibliography}{15}
\expandafter\ifx\csname natexlab\endcsname\relax\def\natexlab#1{#1}\fi

\bibitem[{Anthropic(2024)}]{opus2024}
Anthropic. 2024.
\newblock \href {https://www.anthropic.com/news/claude-3-family} {Claude-opus technical report}.

\bibitem[{Barriere et~al.(2023)Barriere, Sedoc, Tafreshi, and Giorgi}]{barriere-etal-2023-findings}
Valentin Barriere, Jo{\~a}o Sedoc, Shabnam Tafreshi, and Salvatore Giorgi. 2023.
\newblock \href {https://doi.org/10.18653/v1/2023.wassa-1.44} {Findings of {WASSA} 2023 shared task on empathy, emotion and personality detection in conversation and reactions to news articles}.
\newblock In \emph{Proceedings of the 13th Workshop on Computational Approaches to Subjectivity, Sentiment, {\&} Social Media Analysis}, pages 511--525, Toronto, Canada. Association for Computational Linguistics.

\bibitem[{Barriere et~al.(2022)Barriere, Tafreshi, Sedoc, and Alqahtani}]{barriere-etal-2022-wassa}
Valentin Barriere, Shabnam Tafreshi, Jo{\~a}o Sedoc, and Sawsan Alqahtani. 2022.
\newblock \href {https://doi.org/10.18653/v1/2022.wassa-1.20} {{WASSA} 2022 shared task: Predicting empathy, emotion and personality in reaction to news stories}.
\newblock In \emph{Proceedings of the 12th Workshop on Computational Approaches to Subjectivity, Sentiment {\&} Social Media Analysis}, pages 214--227, Dublin, Ireland. Association for Computational Linguistics.

\bibitem[{Chatterjee et~al.(2019)Chatterjee, Narahari, Joshi, and Agrawal}]{chatterjee-etal-2019-semeval}
Ankush Chatterjee, Kedhar~Nath Narahari, Meghana Joshi, and Puneet Agrawal. 2019.
\newblock \href {https://doi.org/10.18653/v1/S19-2005} {{S}em{E}val-2019 task 3: {E}mo{C}ontext contextual emotion detection in text}.
\newblock In \emph{Proceedings of the 13th International Workshop on Semantic Evaluation}, pages 39--48, Minneapolis, Minnesota, USA. Association for Computational Linguistics.

\bibitem[{Conneau et~al.(2020)Conneau, Khandelwal, Goyal, Chaudhary, Wenzek, Guzmán, Grave, Ott, Zettlemoyer, and Stoyanov}]{conneau2020unsupervised}
Alexis Conneau, Kartikay Khandelwal, Naman Goyal, Vishrav Chaudhary, Guillaume Wenzek, Francisco Guzmán, Edouard Grave, Myle Ott, Luke Zettlemoyer, and Veselin Stoyanov. 2020.
\newblock \href {http://arxiv.org/abs/1911.02116} {Unsupervised cross-lingual representation learning at scale}.

\bibitem[{GemmaTeam(2024)}]{gemmateam2024gemma}
GemmaTeam. 2024.
\newblock \href {http://arxiv.org/abs/2403.08295} {Gemma: Open models based on gemini research and technology}.

\bibitem[{Hu et~al.(2021)Hu, Shen, Wallis, Allen-Zhu, Li, Wang, Wang, and Chen}]{hu2021lowrank}
Edward~J. Hu, Yelong Shen, Phillip Wallis, Zeyuan Allen-Zhu, Yuanzhi Li, Shean Wang, Lu~Wang, and Weizhu Chen. 2021.
\newblock \href {http://arxiv.org/pdf/2106.09685} {Lora: Low-rank adaptation of large language models}.

\bibitem[{Izmailov et~al.(2019)Izmailov, Podoprikhin, Garipov, Vetrov, and Wilson}]{izmailov2019averaging}
Pavel Izmailov, Dmitrii Podoprikhin, Timur Garipov, Dmitry Vetrov, and Andrew~Gordon Wilson. 2019.
\newblock \href {http://arxiv.org/abs/1803.05407} {Averaging weights leads to wider optima and better generalization}.

\bibitem[{Jiang et~al.(2023)Jiang, Sablayrolles, Mensch, Bamford, Chaplot, de~las Casas, Bressand, Lengyel, Lample, Saulnier, Lavaud, Lachaux, Stock, Scao, Lavril, Wang, Lacroix, and Sayed}]{jiang2023mistral}
Albert~Q. Jiang, Alexandre Sablayrolles, Arthur Mensch, Chris Bamford, Devendra~Singh Chaplot, Diego de~las Casas, Florian Bressand, Gianna Lengyel, Guillaume Lample, Lucile Saulnier, Lélio~Renard Lavaud, Marie-Anne Lachaux, Pierre Stock, Teven~Le Scao, Thibaut Lavril, Thomas Wang, Timothée Lacroix, and William~El Sayed. 2023.
\newblock \href {http://arxiv.org/abs/2310.06825} {Mistral 7b}.

\bibitem[{Liu et~al.(2019)Liu, Ott, Goyal, Du, Joshi, Chen, Levy, Lewis, Zettlemoyer, and Stoyanov}]{liu2019roberta}
Yinhan Liu, Myle Ott, Naman Goyal, Jingfei Du, Mandar Joshi, Danqi Chen, Omer Levy, Mike Lewis, Luke Zettlemoyer, and Veselin Stoyanov. 2019.
\newblock \href {http://arxiv.org/abs/1907.11692} {Roberta: A robustly optimized bert pretraining approach}.

\bibitem[{Maladry et~al.(2024)Maladry, Singh, and Lefever}]{Maladry2024}
Aaron Maladry, Pranaydeep Singh, and Els Lefever. 2024.
\newblock Findings of the wassa 2024 exalt shared task on explainability for cross-lingual emotion in tweets.
\newblock In \emph{Proceedings of the 14th Workshop of on Computational Approaches to Subjectivity, Sentiment \& Social Media Analysis@ACL 2024}, Bangkok, Thailand.

\bibitem[{Mohammad et~al.(2018)Mohammad, Bravo-Marquez, Salameh, and Kiritchenko}]{mohammad-etal-2018-semeval}
Saif Mohammad, Felipe Bravo-Marquez, Mohammad Salameh, and Svetlana Kiritchenko. 2018.
\newblock \href {https://doi.org/10.18653/v1/S18-1001} {{S}em{E}val-2018 task 1: Affect in tweets}.
\newblock In \emph{Proceedings of the 12th International Workshop on Semantic Evaluation}, pages 1--17, New Orleans, Louisiana. Association for Computational Linguistics.

\bibitem[{OpenAI(2024)}]{openai2024gpt4}
OpenAI. 2024.
\newblock \href {http://arxiv.org/abs/2303.08774} {Gpt-4 technical report}.

\bibitem[{Patkar et~al.(2023)Patkar, Chandrashekhar, and Kadiyala}]{patkar-etal-2023-adityapatkar}
Aditya Patkar, Suraj Chandrashekhar, and Ram Mohan~Rao Kadiyala. 2023.
\newblock \href {https://doi.org/10.18653/v1/2023.wassa-1.46} {{A}ditya{P}atkar at {WASSA} 2023 empathy, emotion, and personality shared task: {R}o{BERT}a-based emotion classification of essays, improving performance on imbalanced data}.
\newblock In \emph{Proceedings of the 13th Workshop on Computational Approaches to Subjectivity, Sentiment, {\&} Social Media Analysis}, Toronto, Canada. Association for Computational Linguistics.

\bibitem[{Touvron et~al.(2023)Touvron, Lavril, Izacard, Martinet, Lachaux, Lacroix, Rozière, Goyal, Hambro, Azhar, Rodriguez, Joulin, Grave, and Lample}]{touvron2023llama}
Hugo Touvron, Thibaut Lavril, Gautier Izacard, Xavier Martinet, Marie-Anne Lachaux, Timothée Lacroix, Baptiste Rozière, Naman Goyal, Eric Hambro, Faisal Azhar, Aurelien Rodriguez, Armand Joulin, Edouard Grave, and Guillaume Lample. 2023.
\newblock \href {http://arxiv.org/abs/2302.13971} {Llama: Open and efficient foundation language models}.

\end{thebibliography}
\bibliographystyle{acl_natbib}

\appendix
\label{Appendix}
\section{Text Translation}
Several translation models and approaches have been tested, with google-translate and utilizing LLMs for translating being the better suited approaches. However the texts were returned without translation in code-mixed text cases is some instances. Despite the higher cost using LLMs worked perfectly in detecting the main language and also to test by translating all texts to English.

\section{Prompts Used}
The prompts used in the system and other analysis tasks were as follows :
\begin{itemize}
    \item \textit{Language Detection of texts} : "Classify given texts as English,Dutch,French,Spanish,Russian. Respond only with one word based on which language the text is in."
    \item \textit{Translation completely to English} : "Translate the text to English. Respond with the same text if already in English completely."
    \item \textit{Classification of Emotion (Proprietary)} : "Classify given texts as Neutral, Joy, Anger, Love, Sadness, Fear. Respond only with one word based on which would be closest classification of user emotion from the text."
    \item \textit{Classification of Emotion (fine-tuned)} : "Given the input text , classify it based on what emotion is being exibited among the following : Joy/Neutral/Anger/Love/Sadness/Fear. Respond with only one emotion only among the options given. Respond with only one word and nothing else."
    \item \textit{Binary Classification of each class separately} : "Is the text indicating \texttt{Class-name} ? Respond only with one word (YES / NO) based on input text."
\end{itemize}

\section{Hyperparameters Used}
Among the hyperparameter space explored for each approach, the best results were obtained with the following values. Rest of the parameters were unspecified during training and hence the default values have been used.

\begin{table}[h]
    \centering
    \begin{tabular}{|c|c|c|}
    \hline
         & \textbf{Transformers} & \textbf{LLMs} \\
    \hline
    No.of Epochs         & 10       & 5       \\
    Learning rate        & 2e-5     & 2e-4    \\
    Weight Decay         & 0.05     & 0.01    \\
    \hline
    \end{tabular}
    \caption{Hyperparameters used in training the models used for fine-tuned transformers and LLMs}
    \label{table:B}
\end{table}

\section{System Replication Instructions}
The system can be replicated using the hyperparameters mentioned in \autoref{table:B} with seed value of 1024. The models used are available on huggingface in various configurations i.e LoRA adapters, 16bit and 4bit precision models. \\

LLAMA-3-8B model trained with additional data from previous years workshop datasets : \\
\begin{itemize}
\item{\href{https://huggingface.co/1024m/EXALT-1A-LLAMA3-5C-Lora}{1024m/EXALT-1A-LLAMA3-5C-Lora}}\\
\item{\href{https://huggingface.co/1024m/EXALT-1A-LLAMA3-5C-16bit}{1024m/EXALT-1A-LLAMA3-5C-16bit}}\\
\item{\href{https://huggingface.co/1024m/EXALT-1A-LLAMA3-5C-4bit}{1024m/EXALT-1A-LLAMA3-5C-4bit}}\\
\end{itemize}

LLAMA-3-8B model trained with datasets translated to English using GPT-4 : \\
\begin{itemize}
\item{\href{https://huggingface.co/1024m/EXALT-1A-LLAMA3-5B-Lora}{1024m/EXALT-1A-LLAMA3-5B-Lora}}\\
\item{\href{https://huggingface.co/1024m/EXALT-1A-LLAMA3-5B-16bit}{1024m/EXALT-1A-LLAMA3-5B-16bit}}\\
\item{\href{https://huggingface.co/1024m/EXALT-1A-LLAMA3-5B-4bit}{1024m/EXALT-1A-LLAMA3-5B-4bit}}\\
\end{itemize}

LLAMA-3-8B model used in the system (main) : \\
\begin{itemize}
\item{\href{https://huggingface.co/1024m/EXALT-1A-LLAMA3-5A-Lora}{1024m/EXALT-1A-LLAMA3-5A-Lora}}\\
\item{\href{https://huggingface.co/1024m/EXALT-1A-LLAMA3-5A-16bit}{1024m/EXALT-1A-LLAMA3-5A-16bit}}\\
\item{\href{https://huggingface.co/1024m/EXALT-1A-LLAMA3-5A-4bit}{1024m/EXALT-1A-LLAMA3-5A-4bit}}\\
\end{itemize}

GEMMA-8B model used in the system (main) : \\
\begin{itemize}
\item{\href{https://huggingface.co/1024m/EXALT-1A-GEMMA-5A-Lora}{1024m/EXALT-1A-GEMMA-5A-Lora}}\\
\item{\href{https://huggingface.co/1024m/EXALT-1A-GEMMA-5A-16bit}{1024m/EXALT-1A-GEMMA-5A-16bit}}\\
\item{\href{https://huggingface.co/1024m/EXALT-1A-GEMMA-5A-4bit}{1024m/EXALT-1A-GEMMA-5A-4bit}}\\
\end{itemize}

Mistral-7B model used in the system (main) : \\
\begin{itemize}
\item{\href{https://huggingface.co/1024m/EXALT-1A-MISTRAL-5A-Lora}{1024m/EXALT-1A-MISTRAL-5A-Lora}}\\
\item{\href{https://huggingface.co/1024m/EXALT-1A-MISTRAL-5A-16bit}{1024m/EXALT-1A-MISTRAL-5A-16bit}}\\
\item{\href{https://huggingface.co/1024m/EXALT-1A-MISTRAL-5A-4bit}{1024m/EXALT-1A-MISTRAL-5A-4bit}}\\
\end{itemize}

\end{document}